\UseRawInputEncoding
\documentclass[letterpaper, 10 pt, conference]{ieeeconf}  % Comment this line out if you need a4paper

\IEEEoverridecommandlockouts                              % This command is only needed if 
\usepackage{graphics} % for pdf, bitmapped graphics files
\usepackage{epsfig} % for postscript graphics files
\usepackage{mathptmx} % assumes new font selection scheme installed
\usepackage{times} % assumes new font selection scheme installed
\usepackage{amsmath} % assumes amsmath package installed
\usepackage{amssymb}  % assumes amsmath package installed
\usepackage{cite}
\usepackage{algorithm}
\usepackage{algorithmic}
\usepackage{subcaption}
 
\usepackage{siunitx}
\usepackage{booktabs}
\usepackage{multirow}
\usepackage{stfloats}
\usepackage{bm}      
\usepackage{bbm}
\usepackage{tabularx} 
\usepackage{xcolor} 
\definecolor{purple}{RGB}{128,0,128} % ????

\title{\LARGE \bf
A New Clustering-based View Planning Method for Building Inspection with Drone
}

\author{Yongshuai Zheng$^{1}$, Guoliang Liu$^{2, *}$, Yan Ding$^{3}$, Guohui Tian$^{4}$% <-this % stops a space
\thanks{This work is partially supported by the Jinan Science and Technology
	Bureau (2021GXRC026), Young Scholars Program of Shandong University
	(2018WLJH71), the Fundamental Research Funds of Shandong University,
	and the Taishan Scholar Foundation of Shandong Province. (Corresponding author: Guoliang Liu.)}% <-this % stops a space
\thanks{The authors are with the School of Control Science and Engineering, Shandong University, Jinan 250061, China (e-mail: liuguoliang@sdu.edu.cn (G. Liu); 202234958@mail.sdu.edu.cn (Y. Zheng); dingyan\_dy@mail.sdu.edu.cn (Y. Ding); g.h.tian@sdu.edu.cn (G. Tian)).
       }%
}

\begin{document}

\maketitle
\thispagestyle{empty}
\pagestyle{empty}

%%%%%%%%%%%%%%%%%%%%%%%%%%%%%%%%%%%%%%%%%%%%%%%%%%%%%%%%%%%%%%%%%%%%%%%%%%%%%%%%
\begin{abstract}

With the rapid development of drone technology, the application of drones equipped with visual sensors for building inspection and surveillance has attracted much attention. View planning aims to find a set of near-optimal viewpoints for vision-related tasks to achieve the vision coverage goal. This paper proposes a new clustering-based two-step computational method using spectral clustering, local potential field method, and hyper-heuristic algorithm to find near-optimal views to cover the target building surface. In the first step, the proposed method generates candidate viewpoints based on spectral clustering and corrects the positions of candidate viewpoints based on our newly proposed local potential field method. In the second step, the optimization problem is converted into a Set Covering Problem (SCP), and the optimal viewpoint subset is solved using our proposed hyper-heuristic algorithm. Experimental results show that the proposed method is able to obtain better solutions with fewer viewpoints and higher coverage.

\end{abstract}


%%%%%%%%%%%%%%%%%%%%%%%%%%%%%%%%%%%%%%%%%%%%%%%%%%%%%%%%%%%%%%%%%%%%%%%%%%%%%%%%
\section{Introduction}

With the rapid advancement of drone technology, the application of drones equipped with visual sensors for surface detection and surveillance of buildings and other large outdoor objects is gaining increasing attention \cite{jing2016sampling, TAN2021103881, liu2024uav}. To realize such applications, it is necessary to solve the view planning problem, which is also called sensor planning in some literatures \cite{9134859, koutecky2016sensor}. The objective of view planning is to utilize visual sensors mounted on robots to find a set of viewpoints \cite{scott2003view, chen2011active}, including their positions and directions, to cover the required surface area of the target object. Common optimization objectives for this problem typically involve improving the surface coverage rate of the target area and reducing the number of viewpoints \cite{jing2018model}.

View planning methods are divided into model-based and non-model-based approaches, depending on on whether the geometric model of the target is known \cite{maboudi2023review}. Model-based methods require a CAD model or existing 3D structural data of the target, which is then converted into triangular mesh models. The level of detail of the mesh directly affects the complexity and accuracy of subsequent processing. In contrast, non-model-based methods operate without prior knowledge of the object \cite{9695293}.
A priori calculation based on the model can ensure sufficient visual coverage, so this paper focuses on the model-based view planning method, especially the three-dimensional visual coverage view planning using drones. Currently, the ``generate-test" approach is a relatively mature technique for view planning. This method is a two-step computational method, where a large number of candidate viewpoints are first generated, and then the problem is transformed into a combinatorial optimization problem based on the constraints of the visual detection task. Subsequently, the optimal subset of viewpoints can be selected using combinatorial optimization methods \cite{scott2009model}. 

In this paper, we propose a new clustering-based two-step computational method for the model-based view planning problem, which is applied to surface detection and surveillance of large buildings. In our method, the input is the triangular mesh model of the target object, and spectral clustering algorithm is utilized to cluster the triangular meshes in the model based on weighted distances and normal vectors. From the clustering results, a set of high-quality candidate viewpoints is generated using the cluster centers and resultant vectors, and a local potential field method is proposed to correct candidate viewpoint positions, thereby achieving higher coverage with fewer viewpoints. In the subsequent steps, the transformed Set Covering Problem is solved using our proposed hyper-heuristic algorithm. The main contributions of this paper are as follows:

\begin{itemize}
	
	\item A new method based on spectral clustering is proposed to generate candidate viewpoints. This method is robust to non-convex cluster shapes, better considers the complex geometric features of objects, and generates high-quality candidate viewpoints.
	\item A new local potential field method is proposed to correct the positions of candidate viewpoints. The positions of candidate viewpoints generated by spectral clustering are not always reasonable, and the local potential field method effectively solves this problem.
	\item A new hyperheuristic algorithm is proposed to solve combinatorial optimization problems. This method is used to search for the optimal viewpoint subset from candidate viewpoints with redundancy, which is more effective and robust than traditional heuristic methods.
	
\end{itemize}

\section{Previous work}

Since the 1980s, researchers have been attempting to solve the model-based view planning problem using various approximation methods. The search space for viewpoints is large, making the problem NP-hard \cite{jing2018model}. Scott et al. \cite{scott2009model} proposed an offset method to generate candidate viewpoints through offsets from the target object surface, transforming the view planning problem into a Set Coverage Problem. Cui et al. \cite{choi2018three} proposed another offset method, designing a normal vector offset and a vertical offset based on the direction of the viewing angle. However, this method generates a large number of viewpoints, incurs high computational costs, and may result in low coverage of large target objects, making it unsuitable for building surface detection.

Jing et al. \cite{jing2016sampling} proposed a two-step iterative random sampling method for view planning that begins with randomly sampling around the target, followed by using a greedy algorithm to address the Set Covering Problem. Although effective, it does not consider the structural characteristics of the target, and its randomness makes the coverage efficiency not high enough. Wang et al. \cite{wang2023high} proposed a parallel deep reinforcement learning approach for industrial inspection, utilizing the Fibonacci numerical integration method for viewpoint sampling within a sphere. This method overcomes the local unevenness seen in traditional approaches but does not adapt well to irregularly shaped object surfaces.

Combinatorial optimization is an important aspect of viewpoint planning. Common optimization methods include heuristic approaches such as ant colony algorithms and genetic algorithms \cite{weerasena2022design}. Hyper-heuristics are a method to enhance the generality of search and optimization algorithms. Unlike traditional heuristics, they consist of high-level and low-level heuristics. High-level heuristics are used to select the most suitable low-level heuristic operators to search for solutions, rather than directly traversing the search space, making them more robust and efficient \cite{dokeroglu2023hyper}. 

Rivera et al. \cite{rivera2023aco} proposed an ACO-based hyper-heuristic for solving multi-robot task allocation problems, using ACO as a high-level strategy and designing several low-level heuristic operators to search for solutions. Zhang et al. \cite{zhang2021multitask} proposed a genetic-based hyper-heuristic algorithm to deal with multi-task scheduling problems, which is more effective and robust than traditional meta-heuristic algorithms. Ferreira et al. \cite{ferreira2015ant} introduced an ACO-based hyper-heuristic for the Set Covering Problem and compared it with the traditional ACO algorithm, demonstrating the effectiveness of hyper-heuristic algorithms.

\begin{figure}[h]
	\centering
	\includegraphics[width=0.98\linewidth]{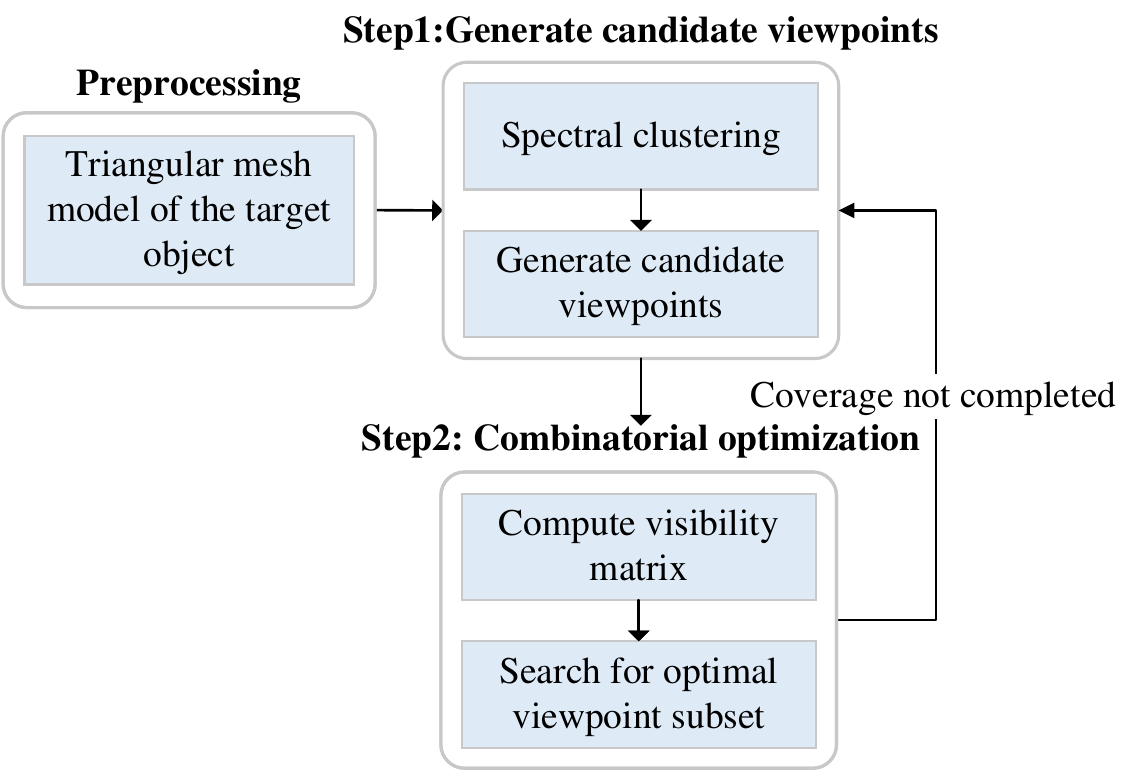}
	\caption{The proposed view planning method.}
	\label{fig:two_step_computational}
\end{figure}

\section{Proposed method}

Our paper proposes a new clustering-based view planning method, which leverages the two-step ``generate-test" computational framework, as shown in Fig. \ref{fig:two_step_computational}. In the first step, candidate viewpoints are generated based on spectral clustering, and the positions of the candidate viewpoints are corrected based on our proposed local potential field method. In the second step, a visibility matrix is calculated, the problem is transformed into a combinatorial optimization problem, and the optimal viewpoint subset is searched based on hyper-heuristic algorithm. There is an iterative process, where if the candidate viewpoints are insufficient to achieve visual coverage of the target object, then the uncovered parts are re-clustered to generate new candidate viewpoints.

\subsection{Generating candidate viewpoints based on spectral clustering}

The purpose of this section is to generate an efficient set of candidate viewpoints from three-dimensional space using spectral clustering algorithm and local potential field method, aiming to narrow down the search space of the solution. The spectral clustering-based method is robust to non-convex clusters, suitable for complex structured datasets \cite{ng2001spectral}, and better considers the geometric features of the target object, generating candidate viewpoints with higher coverage efficiency. However, the positions of candidate viewpoints generated based on clustering algorithms are not always reasonable. The proposed local potential field method effectively corrects the positions of viewpoints.

\subsubsection{Spectral clustering algorithm}

\begin{figure}[h]
	\centering
	\begin{subfigure}{0.42\linewidth}
		\includegraphics[width=\linewidth]{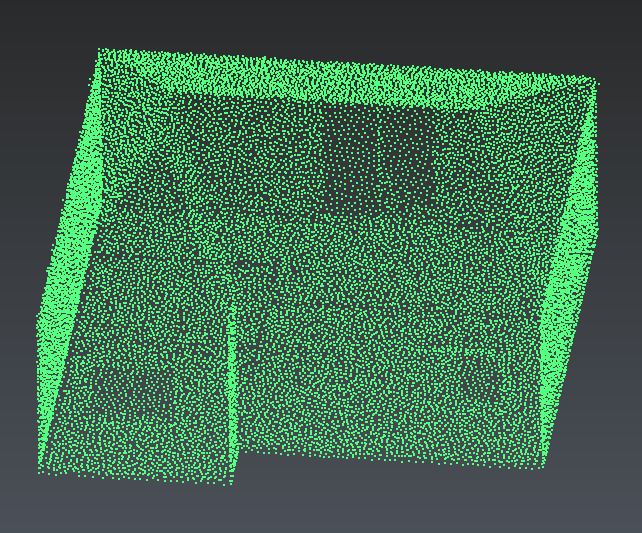}
		\caption{Point cloud model of the target object}
		\label{fig:house_point}
	\end{subfigure}%
	\hspace{0.15cm}
	\begin{subfigure}{0.42\linewidth}
		\includegraphics[width=\linewidth]{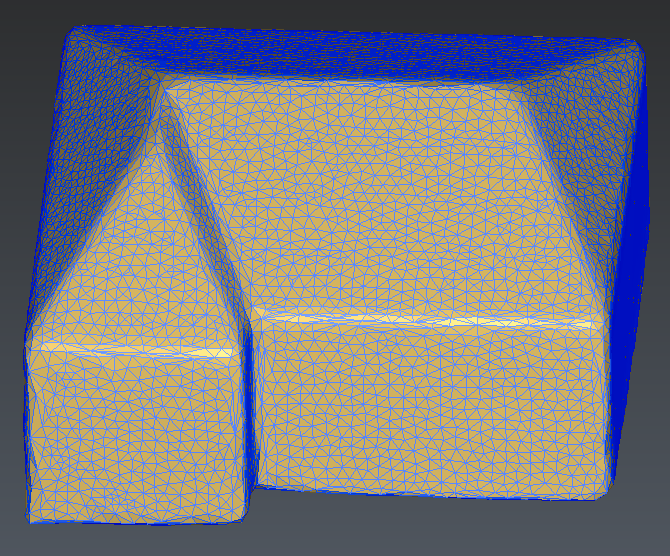}
		\caption{Subdivided triangular mesh model}
		\label{fig:house_mesh}
	\end{subfigure}
	\caption{Processing the target model as a set of triangular meshes.}
	\label{fig:house_model}
\end{figure}

Spectral clustering is a method that utilizes graph and spectral theory by treating the dataset as graph nodes \cite{he2022diagnosis}. In this study, we use spectral clustering to group the triangular meshes in the target object's mesh model, based on the distances and angles between their normal vectors, to generate candidate viewpoints. Before clustering, the target model is preprocessed into a subdivided triangular mesh model, as shown in Fig. \ref{fig:house_model}. Here is an overview of the spectral clustering algorithm design.

First, we compute the adjacency matrix $\bm{G}$. We have adjusted the adjacency matrix calculation by considering both the distances between triangular meshes and the angles between their normal vectors. Here is the formula for $\bm{G}$:
\begin{equation}
	G_{i j} = \theta^* \mathrm{~s}_{i j} + (1-\theta)^* \gamma_{i j} \label{eq:Adjacency matrix calculation}
\end{equation}
where $\theta \in(0,1)$, denotes the weight assigned to the distance and the angle between the normal vectors, $G_{i j}$ denotes the cost of clustering the $i^{th}$ and $j^{th}$ triangular meshes together, $\mathrm{~s}_{i j}$ denotes the distance between the centers of the triangular meshes, and $\gamma_{i j}$ denotes the angle between the normal vectors of the triangular meshes. 

Second, we construct the similarity matrix $\bm{W}$, where the geometric significance of the similarity matrix is to calculate the similarity between triangular meshes. In this paper, we construct a fully connected similarity matrix $\bm{W}$ using a Gaussian kernel function \cite{ng2001spectral}, as follows:
\begin{equation}
	W_{i j} = \exp \left(-\frac{G_{\mathrm{ij}}^2}{2 \sigma^2}\right) \label{eq:Similarity matrix calculation}
\end{equation}
where $W_{i j}$ denotes the similarity between triangular meshes $i^{th}$ and $j^{th}$, and $\sigma$ is a parameter of the Gaussian kernel function used to control the rate of similarity decay. The lower the similarity, the more expensive it is to cluster together and the less likely it is to cluster into a cluster and vice versa.

Third, we compute the laplacian matrix $\bm{L}_{\mathrm{rw}}$ using random walk normalization \cite{shi2000normalized} to enhance algorithm stability:
\begin{equation}
	\bm{L}_{\mathrm{rw}} = \bm{I} - \bm{D}^{-1} \bm{W} \label{eq:Laplace_matrix_calculation}
\end{equation}
where $\bm{I}$ is the unit matrix, $\bm{D}$ is the degree matrix, and the $i^{th}$ diagonal element of $\bm{D}$ is the degree of data point $i$, denoted as $D_{i j}=\sum_{\mathrm{j}} W_{i j}$.

Subsequently, we perform eigenvalue decomposition on the normalized Laplacian matrix $\bm{L}_{\mathrm{rw}}$ to extract eigenvalues and select eigenvectors for the top $k$ non-zero eigenvalues, where $k$ is the number of clusters. These eigenvectors are then used to create a new data representation for K-means clustering, effectively grouping the triangular meshes into different clusters.

\begin{figure}[h]
	\centering
	\includegraphics[width=0.53\linewidth]{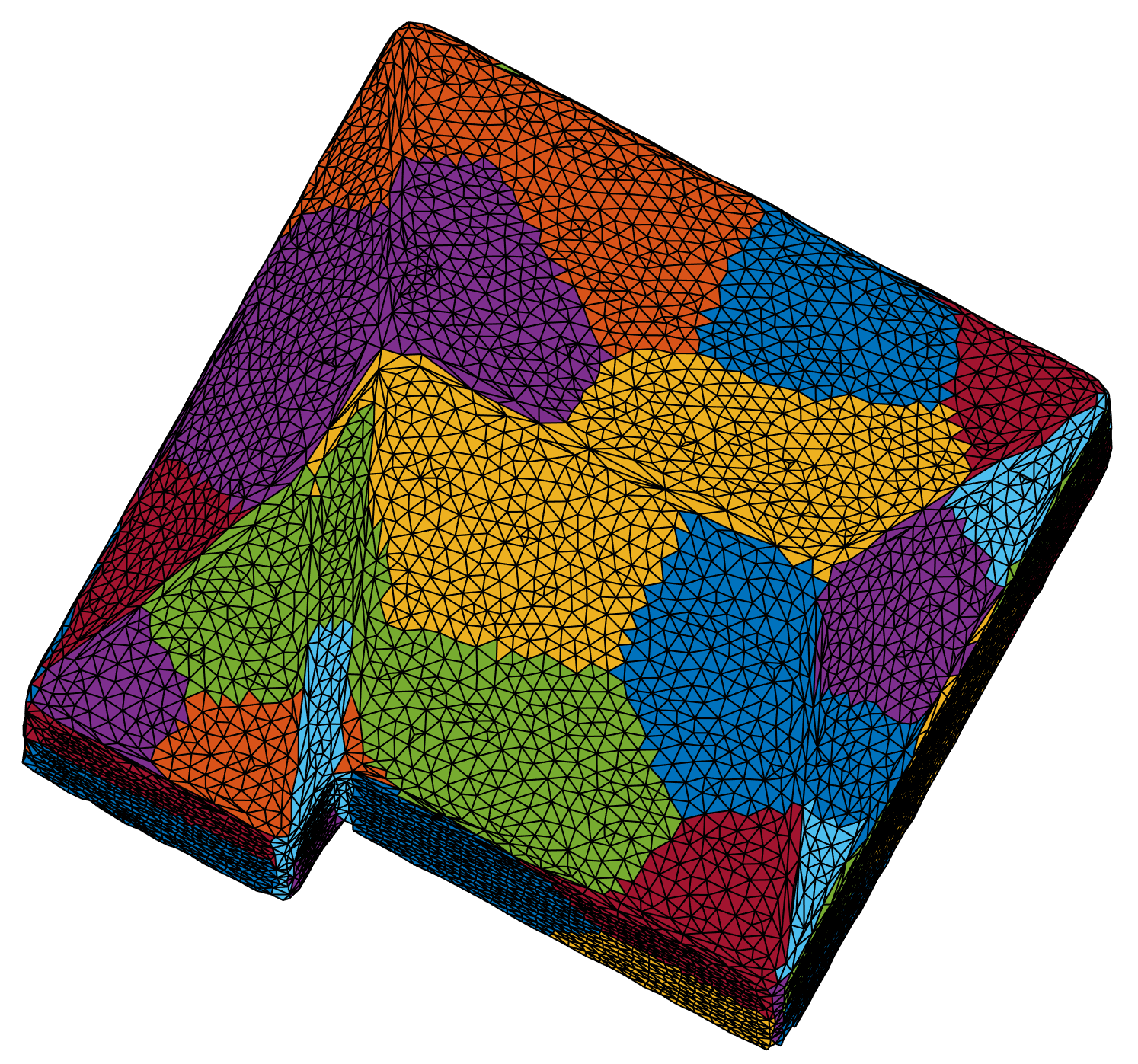}
	\caption{Spectral clustering of a set of triangular meshes.}
	\label{fig:cluster}
\end{figure}

\subsubsection{Generate candidate viewpoints}

\begin{figure}[h]
	\centering
	\includegraphics[width=0.6\linewidth]{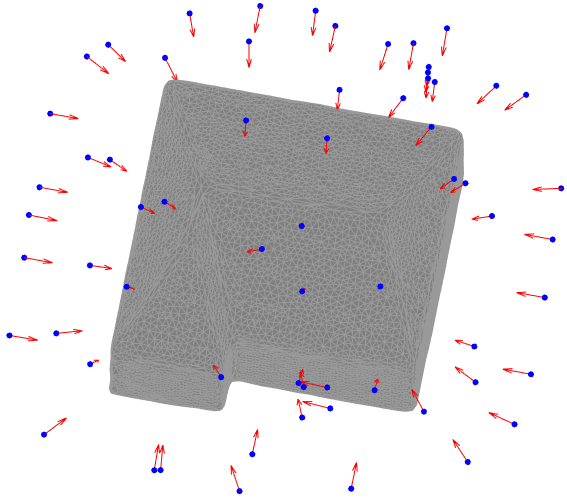}
	\caption{Generated candidate viewpoints.}
	\label{fig:candidate_vps_patches}
\end{figure}

After clustering, we compute the centers of all clusters and the resultant vector of the normals of all triangular meshes within each cluster. Then, we translate the coordinates of the cluster centers along the direction of the resultant vector, thereby obtaining the positions of candidate viewpoints. The opposite of the resultant vector is used as the direction of the viewpoints. The generation of candidate viewpoints is shown in Fig. \ref{fig:candidate_vps_patches}, and the specific calculation process is derived as follows.

We first compute the clustering centers $\bm{C}$ and resultant vectors $\bm{N}$:
\begin{equation}
	\bm{C}_k = \frac{1}{t} \cdot \sum_{i=1}^t \bm{x}_i \label{eq:cluster_center}
\end{equation}
\begin{equation}
	\bm{N}_k = \frac{1}{t} \cdot \sum_{i=1}^t \frac{\bm{v}_i}{\left\|\bm{v}_i\right\|} \label{eq:resultant_vector}
\end{equation}
where the subscript $k$ denotes the index of the clusters, $t$ denotes that the cluster contains $t$ triangular meshes, $\bm{x}_i$ denotes the centroid of the triangular mesh in the cluster, $\bm{v}_i$ denotes the normal vector of the triangular mesh, $\bm{C}_k$ and $\bm{N}_k$ denote the $k^{th}$ clustering center and the resultant vector, respectively.

We then generate candidate viewpoint positions $\bm{P}$ and directions $\bm{V}$:
\begin{equation}
	\bm{P}_k = \bm{C}_k + d \cdot \bm{N}_k \label{eq:vp_position}
\end{equation}
\begin{equation}
	\bm{V}_k = \bm{C}_k - \bm{P}_k \label{eq:vp_direction}
\end{equation}
where $\bm{P}_k$ denotes the position of the $k^{th}$ candidate viewpoint, $\bm{V}_k$ denotes the direction of the $k^{th}$ candidate viewpoint, and $d$ denotes the distance that the clustering center moves along the concatenated vector. It has been experimentally tested that if the Field of Depth (FOD) of the camera is known, then by setting $d$ to about 0.95 times the FOD generates relatively higher quality viewpoints.

Additionally, it is crucial to define the constraint space for generating candidate viewpoints to ensure their positions meet specific requirements. We propose a local potential field method to correct the positions of viewpoints, similar to the calculation of repulsive and attractive forces in artificial potential fields \cite{tian2021overall}. We set constraints on the positions of candidate viewpoints: (1) Distance limitations to prevent the drone from being too close to the target structure; (2) Altitude limitations to ensure the drone's flight altitude is not too low, avoiding proximity to the ground:
\begin{equation}
	\begin{aligned}
		\bm{v}_{\text{correct}} &= \frac{\sum_{i}^{N}(\bm{P}_{k}-\bm{p}_{\text{mesh}_{i}})}{\left\|\sum_{i}^{N}(\bm{P}_{k}-\bm{p}_{\text{mesh}_{i}})\right\|}, \\
		\text{for all:} & \quad \{\bm{p}_{\text{mesh}_{i}} \mid \left\|\bm{P}_{k}-\bm{p}_{\text{mesh}_{i}}\right\|<d_{\text{safe}}\}
	\end{aligned}
	\label{eq:probabilistic_potential_field_method}
\end{equation}
\begin{equation}
	\bm{v}_{\text{correct}} = \frac{\bm{h}_{\text{limit}} - \bm{P}_k}{\left\|\bm{h}_{\text{limit}} - \bm{P}_k\right\|}, \quad \text{if } \bm{P}_{k_z} < \bm{h}_{\text{limit}_z}
	\label{eq:probabilistic2}
\end{equation}

Eq. \eqref{eq:probabilistic_potential_field_method} computes the correction direction when the candidate viewpoint is too close to the target object by summing the repulsive forces from the triangular meshes located within the distance limit. Eq. \eqref{eq:probabilistic2} computes the correction direction when the candidate viewpoint position is below a certain height by calculating the attraction towards the predefined height. Here, $N$ denotes the number of triangular meshes within the distance limit, $\bm{P}_k$ denotes the position of the candidate viewpoint, $\bm{p}_{\text{mesh}_{i}}$ denotes the position of the $i^{th}$ triangular mesh, $d_{\text{safe}}$ defines the safe distance, $\bm{P}_{k_z}$ and $\bm{h}_{\text{limit}_z}$ respectively represent the height of the candidate viewpoint and the safe height.

Subsequently, the candidate viewpoint position is corrected based on the correction direction $\bm{v}_{\text{correct}}$. If multiple adjustments of the candidate viewpoint position based on the calculated $\bm{v}_{\text{correct}}$ still do not satisfy the constraint conditions, then the distance $d$ moved by the cluster centroid along the resultant vector during the generation of candidate viewpoints is modified based on Eq. \eqref{eq:vp_position} to find the position that satisfies the constraint conditions:
\begin{equation}
	\begin{aligned}
		& \bm{P}_k = \bm{C}_k + \left(d - i \cdot d_{\text{step}}\right) \cdot \bm{N}_k \\
	\end{aligned}
	\label{eq:vp position2}
\end{equation}
where $d_\text{{step}}$ is the distance to the clustering center and $i$ is the number of iterations. After each iteration, the local potential field method is reapplied to adjust the position of the candidate viewpoint. Accordingly, when the candidate viewpoint position is corrected to meet the constraints, Eq. \eqref{eq:vp_direction} is applied to recalculate the direction of the viewpoint.

\subsection{Conversion to a combinatorial optimization problem}

After generating candidate viewpoints, the problem is transformed into a Set Covering Problem \cite{scott2009model} through the calculation of the visibility matrix.

\subsubsection{Computation of the visibility matrix}

\begin{figure}[h]
	\centering
	\begin{subfigure}{0.4\linewidth}
		\includegraphics[width=\linewidth]{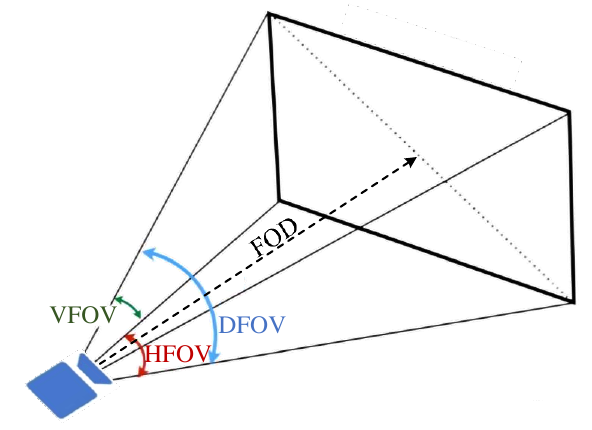}
		\caption{}
		\label{fig:visibility_model_a}
	\end{subfigure}%
	\hspace{0.3cm}
	\begin{subfigure}{0.4\linewidth}
		\includegraphics[width=\linewidth]{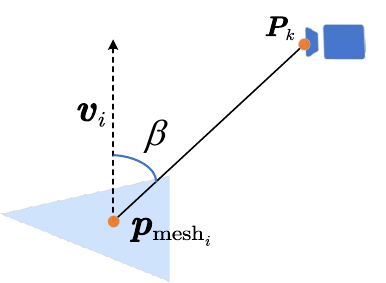}
		\caption{}
		\label{fig:visibility_model_b}
	\end{subfigure}
	\caption{The visibility model.}
	\label{fig:visibility_models}
\end{figure}

\begin{figure}[h]
	\centering
	\begin{subfigure}{0.35\linewidth}
		\includegraphics[width=\linewidth]{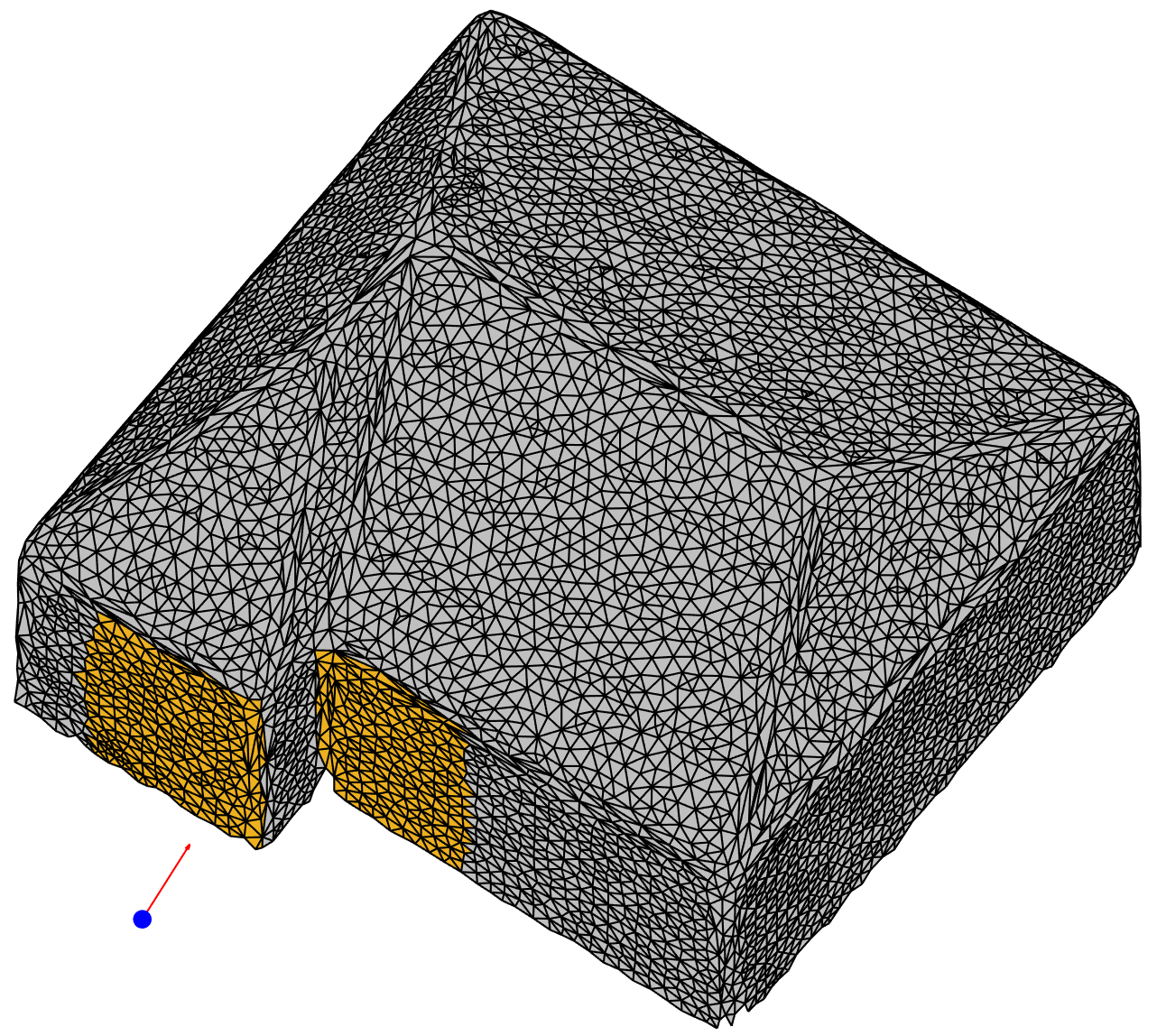}
		\caption{}
		\label{fig:visible_areas1}
	\end{subfigure}%
	\hspace{0.3cm}
	\begin{subfigure}{0.35\linewidth}
		\includegraphics[width=\linewidth]{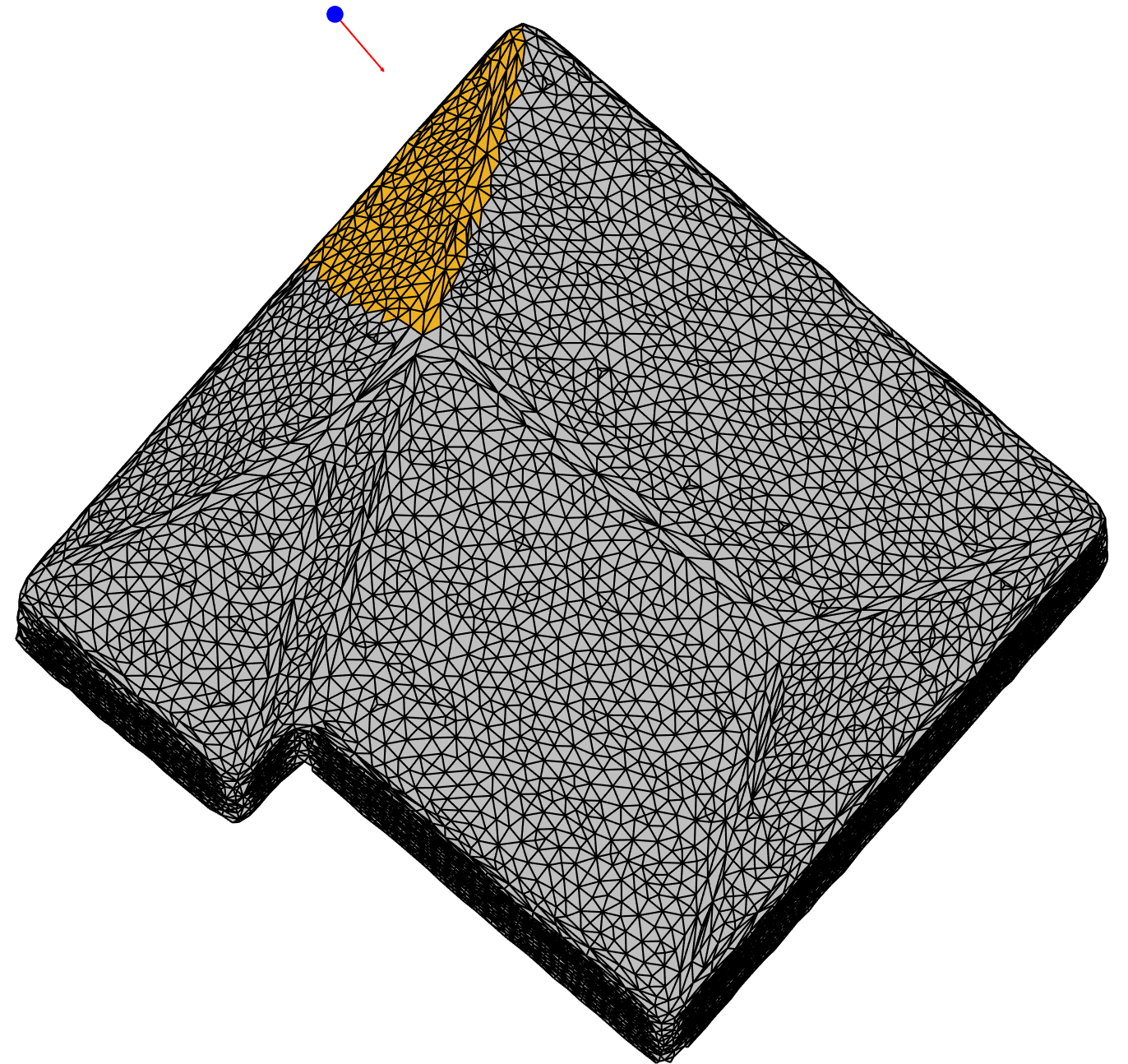}
		\caption{}
		\label{fig:visible_areas2}
	\end{subfigure}
	\caption{Example of visible area from a given viewpoint. Visible areas are shown in yellow.}
	\label{fig:visible_areas}
\end{figure}

The visibility matrix is a binary matrix that indicates which triangular meshes are visible from each viewpoint. It is an $m \times n$ matrix, where $m$ is the number of candidate viewpoints and $n$ is the number of triangular meshes. An element in the matrix is set to 1 if a triangular mesh is visible from a viewpoint and 0 otherwise. Fig. \ref{fig:visible_areas} shows an example of the visible area from a given viewpoint. A triangular mesh is considered visible from a viewpoint if it meets the following conditions:

\begin{itemize}
	\item The triangular mesh must be within the camera's field of depth (FOD) from the viewpoint, as Fig. \ref{fig:visibility_model_a} shows.
	\item The triangular mesh must be within the camera's field of view (FOV) from the viewpoint, as Fig. \ref{fig:visibility_model_a} shows.
	\item The viewing angle $\beta$ must be within a certain range predicted by the camera specifications, as Fig. \ref{fig:visibility_model_b} shows.
	\item There should be no other obstructing structures between the triangular mesh and the viewpoint, which can be determined using ray-triangle intersection tests \cite{moller2005fast}.
\end{itemize}

\subsubsection{Set Covering Problem formulation}

The Set Covering Problem (SCP) is a classical problem in combinatorial optimization, which involves finding a set of subsets to cover all elements at minimum cost. It is a problem known to be NP-hard,  typically solved using heuristic algorithms. In this paper, we reformulate the problem as a constrained combinatorial optimization problem \cite{jing2018model}, where the objective is to find the minimum subset of viewpoints that can cover a certain proportion of triangular meshes:
\begin{equation}
	\begin{aligned}{} 
		& \min \sum\limits_{i=1}^m s_i, \quad \text { where } s_i \in\{0,1\} \\
		\text { s.t. } & \sum\limits_{j=1}^n\left(\left(\sum\limits_{i=1}^m s_i A_{i j}\right) \geqslant 1\right) \geqslant \delta \cdot n
	\end{aligned}
	\label{eq:Formula for Partial Set Covering Problem}
\end{equation}
where $s_i$ denotes the $i^{th}$ candidate viewpoint, with $s_i=1$ indicating selection and $s_i=0$ indicating non-selection; $\bm{A}$ is the $m \times n$ visibility matrix; $\delta$ denotes the preset target coverage ratio, which is the proportion of triangular meshes that can be observed at least once.

\subsection{Genetic-based hyper-heuristic algorithm}

This paper proposes a new genetic-based hyper-heuristic algorithm, GA-HH, which utilizes a genetic algorithm as the high-level heuristic (HLH) and designs seven efficient low-level heuristic operators (LLH) based on the neighborhood structure. {Using the genetic algorithm to manage low-level heuristics enhances robustness and flexibility compared to traditional heuristics. Additionally, the global search capability of the genetic strategy is strong, helping to prevent the algorithm from prematurely converging on local optima.

\begin{figure}[h]
	\centering
	\includegraphics[width=0.7\linewidth]{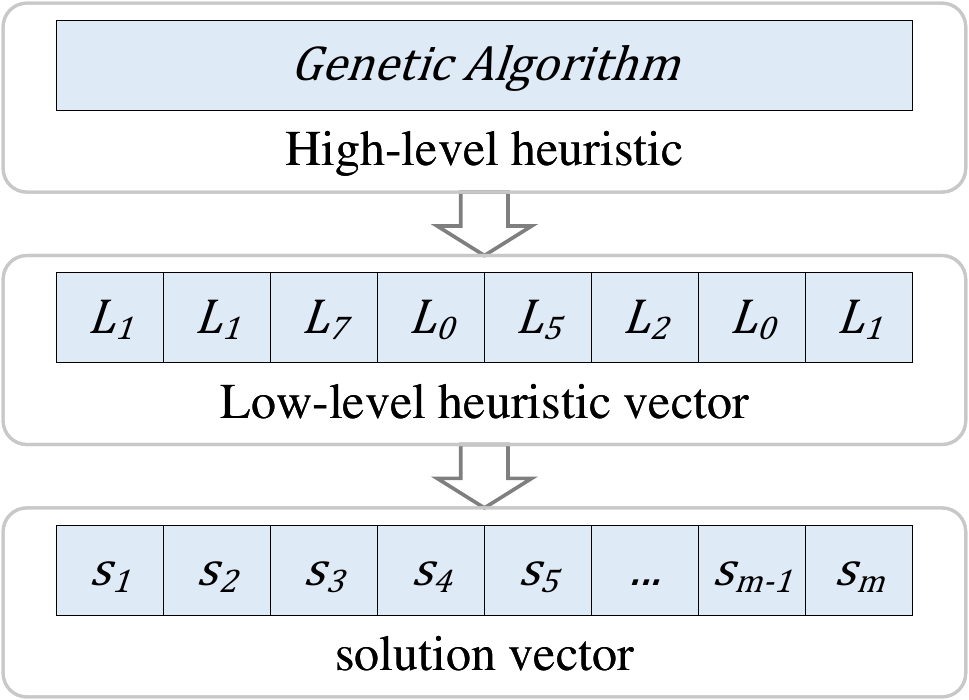}
	\caption{GA-HH: Genetic-based hyper-heuristic algorithm.}
	\label{fig:GA-HH}
\end{figure}

As shown in Fig. \ref{fig:GA-HH}, the HLH does not directly solve the problem but instead manages LLHs to find the most effective heuristic operators for the task. Each element in the low-level heuristic vector represents a chosen operator aimed at improving the fitness of the solution vector. Notably, the label "$L_0$" indicates that no operator is selected for that element. The solution vector $\bm{s}$ is a binary vector, where $s_i$ denotes the $i^{th}$ candidate viewpoint, $s_i=1$ means the viewpoint is selected, and $s_i=0$ means it is not.

\subsubsection{High-level heuristic}

The genetic algorithm primarily consists of several components, including fitness evaluation, initialization, selection, crossover, and mutation \cite{revuelta2021optimization}.

\textit{Initialization:} The initialization process sets up the solution and low-level heuristic vectors. In GA-HH, low-level heuristic vectors are randomly generated, while a greedy search-derived solution serves as the initial solution vector to quicken convergence.

\textit{Fitness Evaluation:} This paper proposes a new fitness calculation method for the Set Covering Problem, enhancing the evolutionary guidance for genetic algorithms. Higher fitness indicates a better solution.

Before assessing fitness, we determine the number of times $\bm{c}$ each triangular mesh is covered and count the triangular meshes $n_{\mathrm{cover}}$ that are covered at least once:
\begin{equation}
	\bm{c} = \bm{s} \cdot \bm{A} \label{eq:covered times}
\end{equation}
\begin{equation}
	n_{\mathrm{cover}} = \sum\limits_{i=1}^n [c_i \geq 1] \label{eq:complete cover}
\end{equation}
where $\bm{s}$ denotes the solution vector, $\bm{A}$ is the visibility matrix, $[c_i \geq 1]$ represents 1 if $c_i \geq 1$, otherwise 0.

The formula for fitness $f$ is:
\begin{equation}
	f =
	\begin{cases}
		(m - \sum_{i=1}^m s_i) + \left(1 - \frac{n_{\text{cover}}}{\sum_{i=1}^n c_i}\right), & \text{if } n_{\text{cover}} \geq \delta \cdot n \\
		\frac{n_{\text{cover}}}{\delta \cdot n}, & \text{if } n_{\text{cover}} < \delta \cdot n
	\end{cases}
	\label{eq:fitness}
\end{equation}
where $m$ is the number of candidate viewpoints. When $n_{\text{cover}} \geq \delta \cdot n$, it means that the number of covered triangular meshes has reached the preset value. The decimal part of fitness, $1-n_{\text{cover}}/\sum_{i=1}^n c_i$, indicates the redundancy of the coverage. A higher redundancy implies that the selected viewpoint subset can cover the triangular meshes more times when the number of selected viewpoints is the same. Consequently, the solution vector is more likely to further reduce the number of selected viewpoints in subsequent evolution.

\textit{Selection:} In genetic algorithms, individuals with higher fitness are more likely to be selected for reproduction. In this stage, roulette wheel selection is used as:

\begin{equation}
	P_i = \frac{f_i}{\sum_{i=1}^k f_i} \label{eq:roulette wheel selection}
\end{equation}
where $f_i$ denotes individual fitness and $k$ denotes population size. During this process, both low-level heuristic vectors and solution vectors need to be replicated synchronously.

\textit{Crossover and Mutation:} During these stages, only low-level heuristic vectors are modified, excluding solution vectors. GA-HH utilizes a segmented crossover strategy by randomly selecting two points and swapping the intervening elements with their neighbors. The mutation involves a ``one-point" approach, randomly mutating one element.

\textit{Generating New Solution Vectors:} After the low-level heuristic vectors are updated through crossover and mutation, they are used to generate new solution vectors. Each element in the low-level heuristic vector represents a low-level heuristic operator, and these operators are used to update the solution vector. If an element is 0, no operator is selected.

\begin{figure*}[ht]
	\centering
	\begin{subfigure}{0.24\textwidth} % Adjust the width as needed
		\centering
		\includegraphics[width=\linewidth]{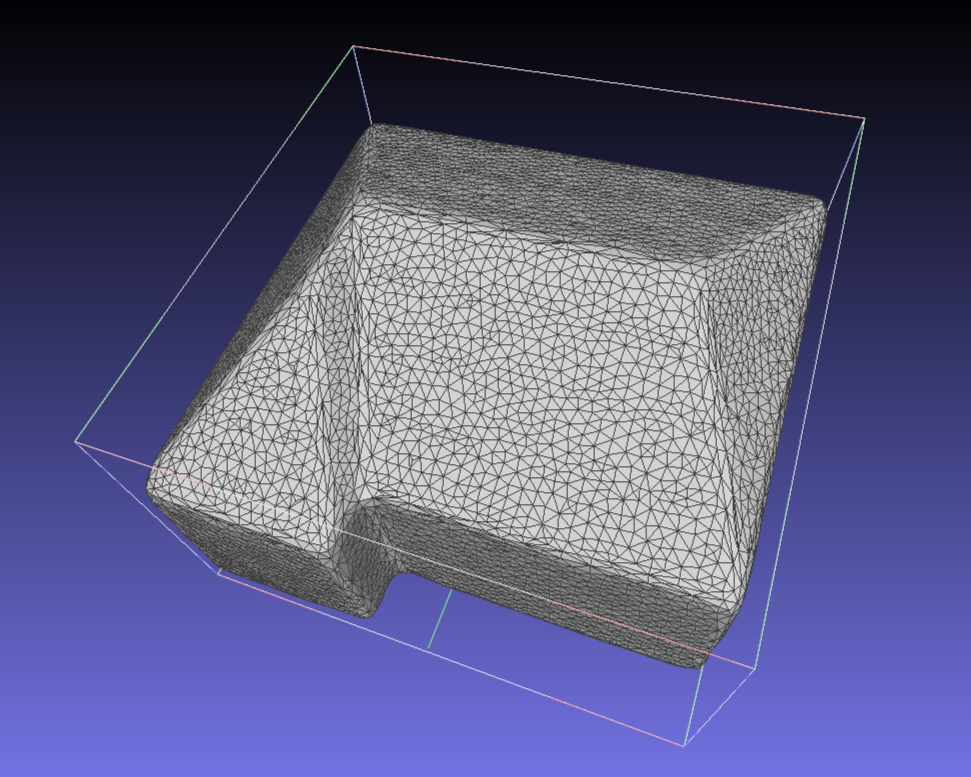}
		\caption{Target building 1: bounding box size is 81 x 81 x 46 meters}
		\label{fig:buliding1}
	\end{subfigure}%
	\hfill
	\begin{subfigure}{0.24\textwidth}
		\centering
		\includegraphics[width=\linewidth]{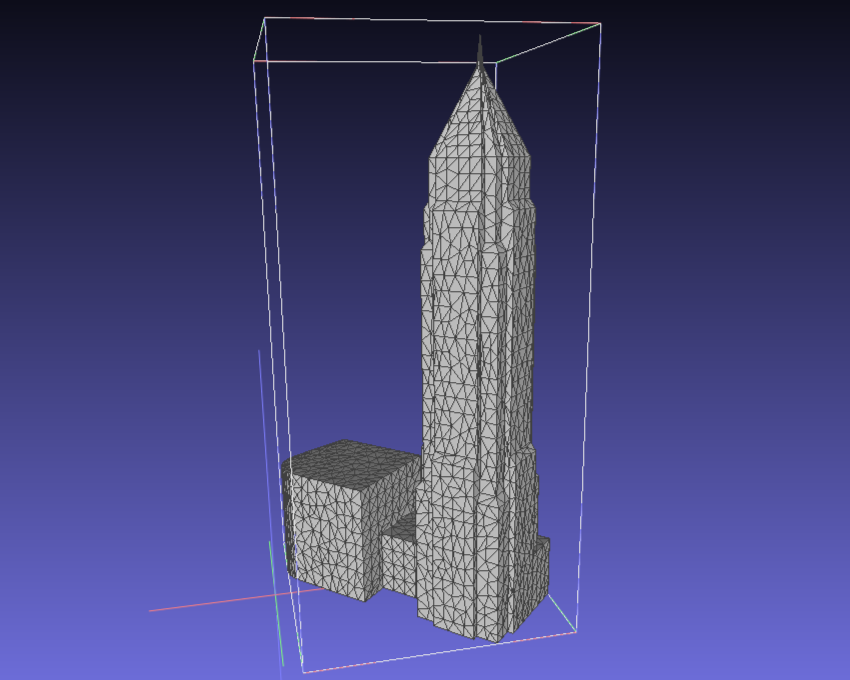}
		\caption{Target building 2: bounding box size is 81 x 82 x 171 meters}
		\label{fig:buliding2}
	\end{subfigure}%
	\hfill
	\begin{subfigure}{0.24\textwidth}
		\centering
		\includegraphics[width=\linewidth]{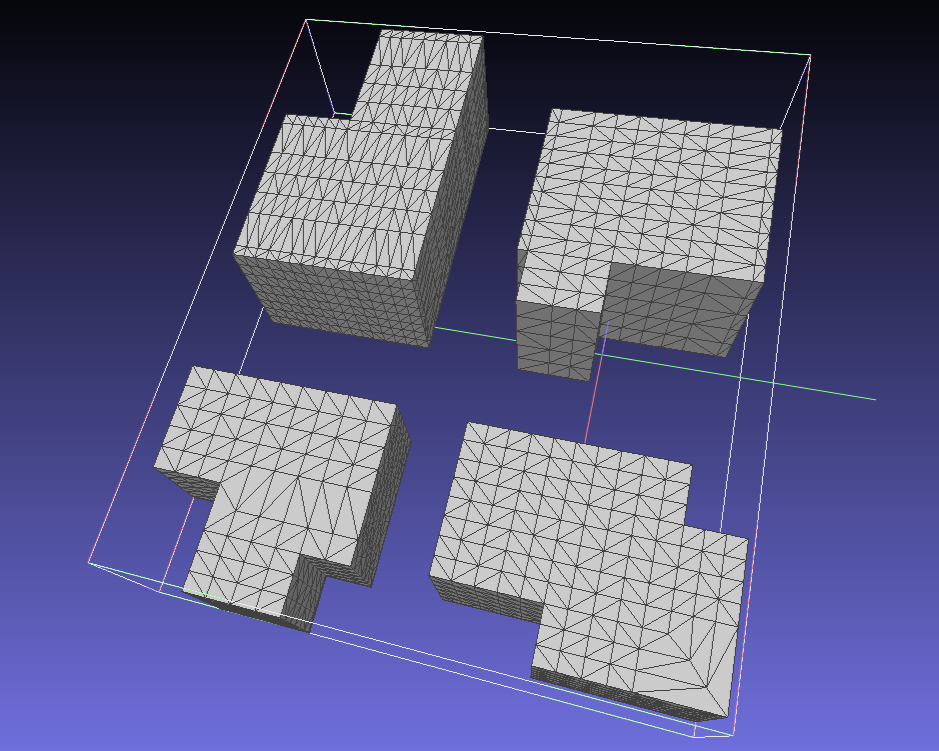}
		\caption{Target building 3: bounding box size is 94 x 77 x 21 meters}
		\label{fig:buliding3}
	\end{subfigure}%
	\hfill
	\begin{subfigure}{0.24\textwidth}
		\centering
		\includegraphics[width=\linewidth]{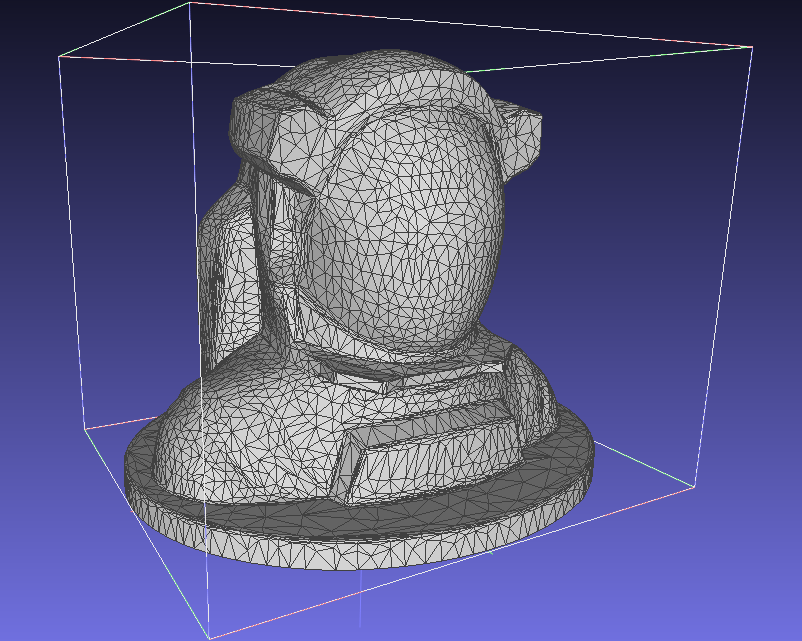}
		\caption{Target building 4: bounding box size is 92 x 77 x 74 meters}
		\label{fig:buliding4}
	\end{subfigure}
	\caption{Triangular mesh model and bounding box of the target building.}
	\label{fig:bulidings}
\end{figure*}

\subsubsection{Low-level heuristics}

Low-level heuristics play a crucial role in bridging the high-level heuristic and the problem solution, and their design significantly impacts the effectiveness of GA-HH. In this paper, seven simple and efficient low-level heuristic operators are designed as follows:

\begin{itemize}
	
	\item LLH1 (random-mutation): 
	This operator randomly selects one element from the solution vector and flips its value. It introduces variability into the solution set and helps prevent the algorithm from stagnating.
	
	\item LLH2 (random-swap): 
	This operator randomly selects two elements from the solution vector and swaps their values, thus enhancing diversity in the solutions by rearranging existing components.
	
	\item LLH3 (section-crossover): 
	For a given solution vector $\bm{S}_1$, this operator selects an adjacent solution vector $\bm{S}_2$, randomly generates two index values, and swaps the element values between these indices in $\bm{S}_1$ and $\bm{S}_2$ to generate a new solution. It combines traits from both vectors, potentially leading to better solutions.
	
	\item LLH4 (adjacent-scattered-crosscover): 
	Inspired by existing research \cite{ferreira2015ant}, this operator takes a solution vector $\bm{S}_1$ and an adjacent vector $\bm{S}_2$ and creates a random bitmask of the same size as the solution. If the value at position $i$ of the mask is 1, the new solution copies the $i^{th}$ value of $\bm{S}_1$(i.e., $\bm{S}_0[i] = \bm{S}_1[i]$), otherwise it copies the $i^{th}$ value of $\bm{S}_2$. It selectively blending their features to form a new solution.
	
	\item LLH5 (random-scattered-crosscover): 
	Similar to LLH4, this operator uses a random bitmask but selects another solution vector $\bm{S}_2$ randomly for the crossover. This increases the diversity by mixing features from potentially non-adjacent solutions, expanding the search space explored by the algorithm.
		
	\item LLH6 (fusion-crosscover): 
	Inspired by literature \cite{wang2020improved}, this operator is similar to LLH4 but adjusts the number of 1s in the random bitmask proportionally to the fitness of the solutions $\bm{S}_1$ and $\bm{S}_2$. This method aims to preserve more advantageous traits from the fitter solution, enhancing the quality of the generated solutions.
	
	\item LLH7 (multi-section-crossover): 
	Extending the concept of LLH3, this operator selects multiple pairs of index values for a given solution vector $\bm{S}_1$ and an adjacent solution vector $\bm{S}_2$ and swaps the element values between these indices. This allows for more complex combinations of their features and increases the potential for creating highly optimized solutions.
	
\end{itemize}

\section{Experimental results}

\begin{figure*}[ht]
	\centering
	\begin{subfigure}{\textwidth}
		\centering
		\includegraphics[width=0.8\textwidth]{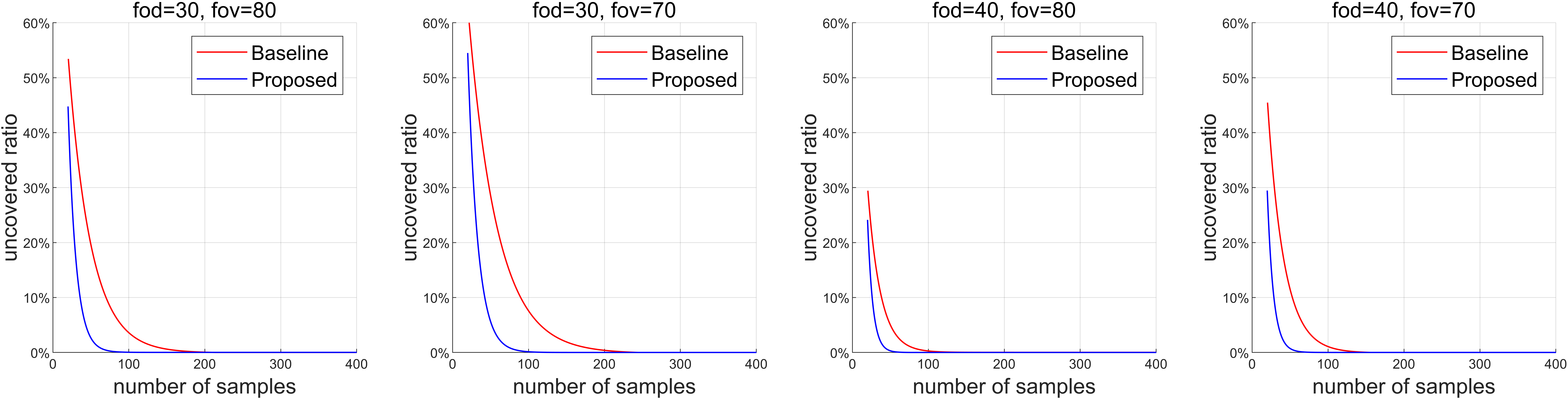}
		\caption{Target building 1}
		\label{fig:result2_1}
	\end{subfigure}%
	
	%\vspace{\baselineskip}
	
	\begin{subfigure}{\textwidth}
		\centering
		\includegraphics[width=0.8\textwidth]{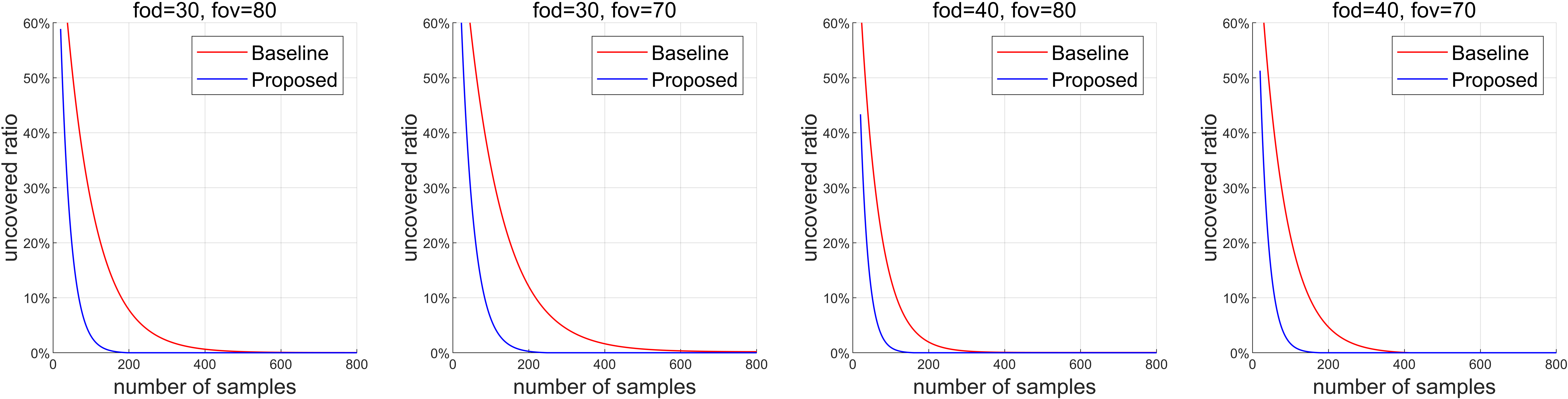}
		\caption{Target building 2}
		\label{fig:result2_2}
	\end{subfigure}
	
	%\vspace{\baselineskip}
	
	\begin{subfigure}{\textwidth}
		\centering
		\includegraphics[width=0.8\textwidth]{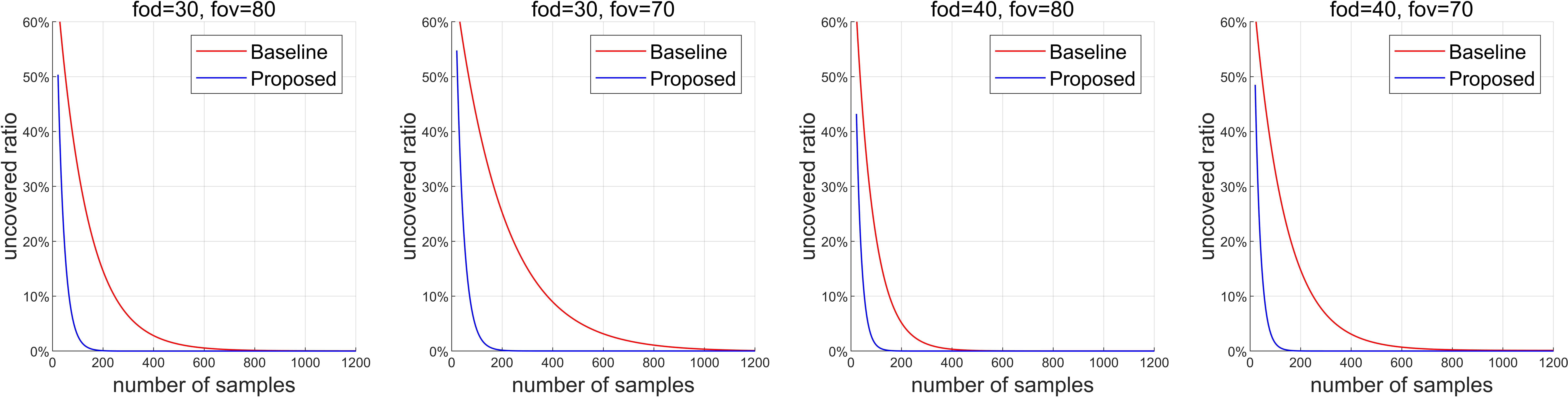}
		\caption{Target building 3}
		\label{fig:result2_3}
	\end{subfigure}
	
	%\vspace{\baselineskip}
	
	\begin{subfigure}{\textwidth}
		\centering
		\includegraphics[width=0.8\textwidth]{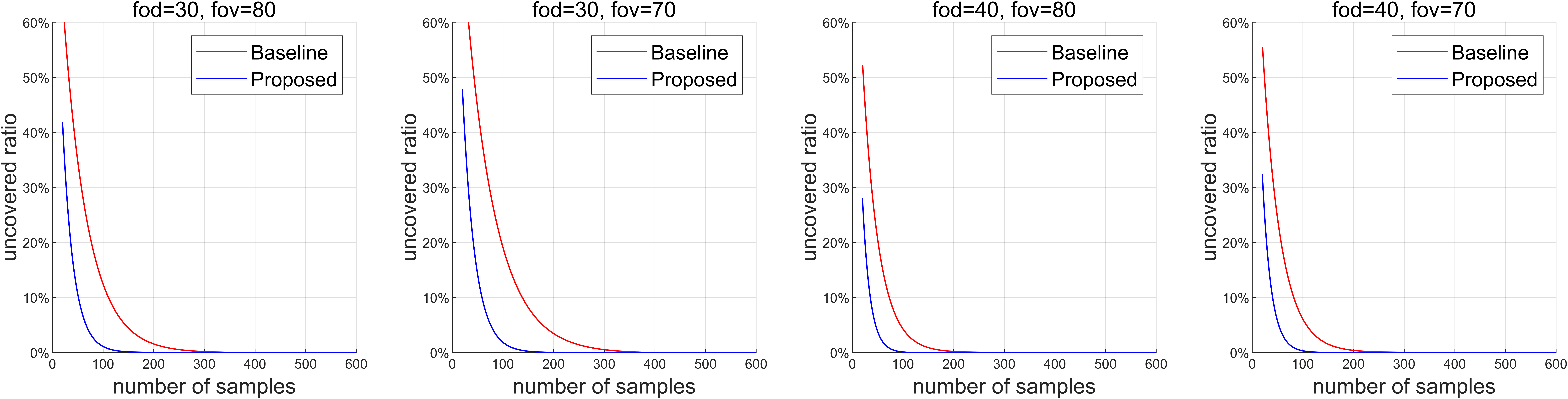}
		\caption{Target building 4}
		\label{fig:result2_4}
	\end{subfigure}

	\caption{Percentage ratios of uncovered areas with different number of generated candidate viewpoints. The proposed method is labeled as ``Proposed" (shown in blue); the method based on random sampling is labeled as ``Baseline" (shown in red). fod is the maximum viewing distance and fov is the maximum viewing angle. The x-axis is the number of generated candidate viewpoints, the y-axis is the non-coverage rate, and the smaller the y-axis value, the better.}
	\label{fig:result2}
\end{figure*}

\begin{figure*}[htpb]
	\centering
	\begin{subfigure}{0.245\textwidth} % Adjust the width as needed
		\centering
		\includegraphics[width=\textwidth]{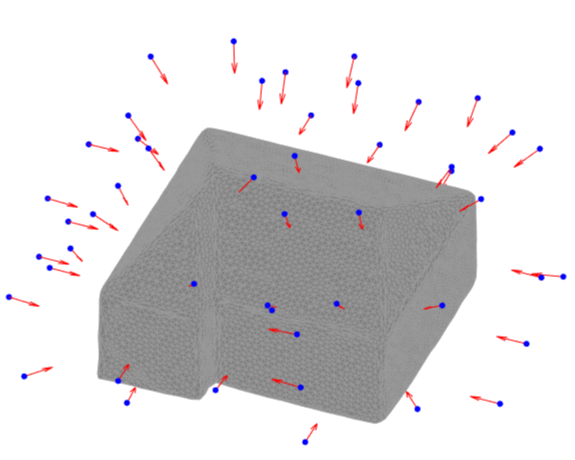}
		\caption{Target building 1}
		\label{fig:buliding1_res}
	\end{subfigure}%
	\hfill
	\begin{subfigure}{0.245\textwidth}
		\centering
		\includegraphics[width=\textwidth]{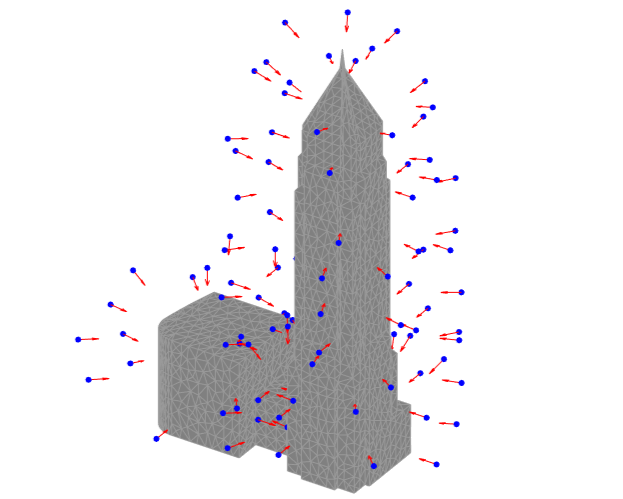}
		\caption{Target building 2}
		\label{fig:buliding2_res}
	\end{subfigure}%
	\hfill
	\begin{subfigure}{0.245\textwidth}
		\centering
		\includegraphics[width=\textwidth]{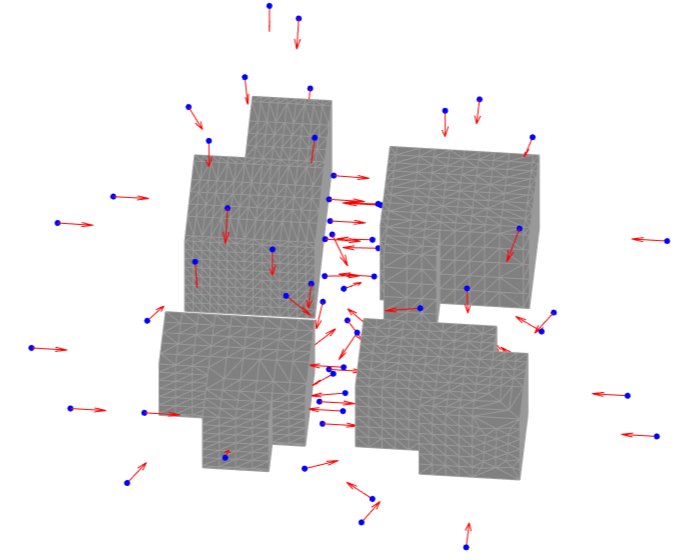}
		\caption{Target building 3}
		\label{fig:buliding3_res}
	\end{subfigure}
	\hfill
	\begin{subfigure}{0.245\textwidth}
		\centering
		\includegraphics[width=\textwidth]{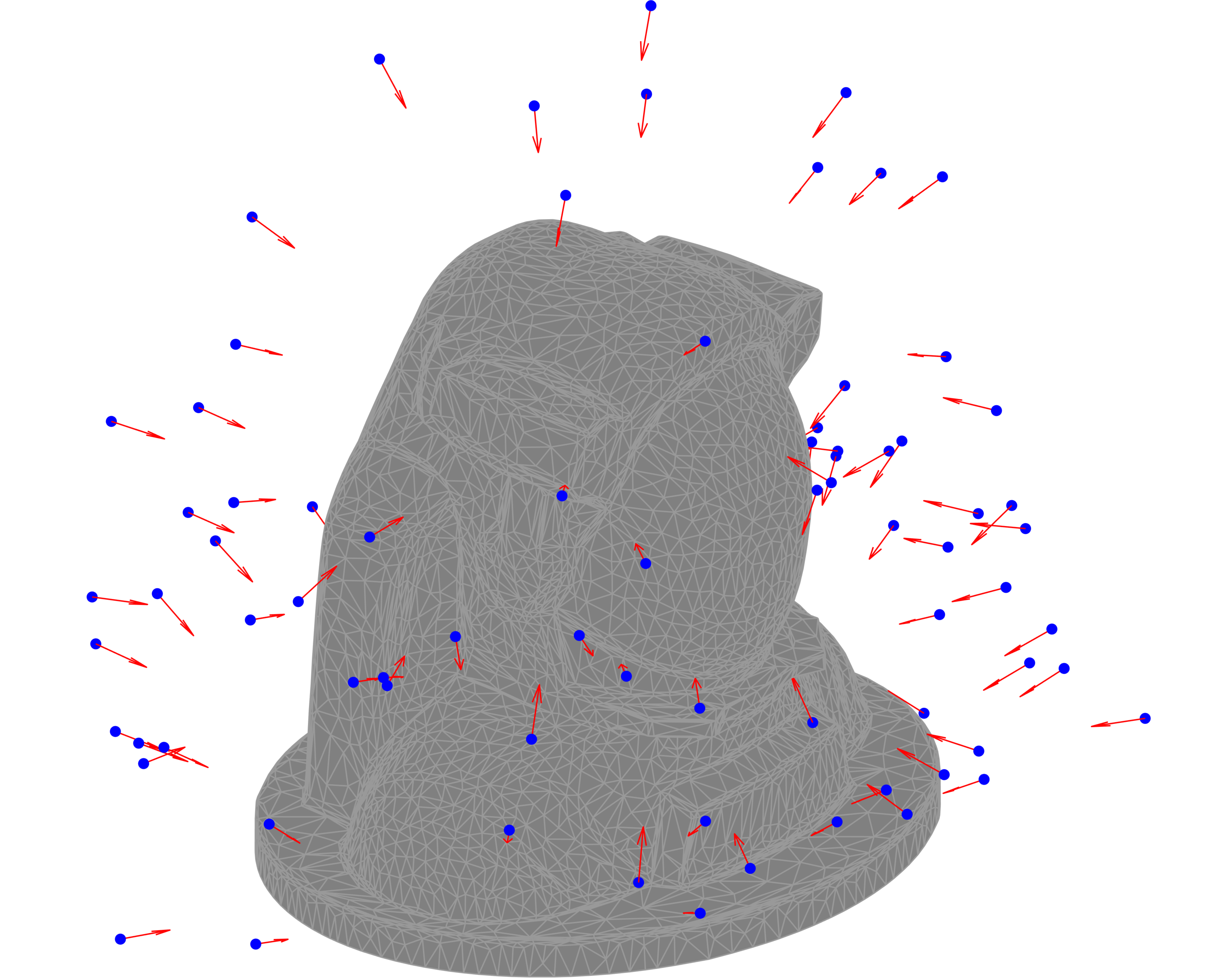}
		\caption{Target building 4}
		\label{fig:buliding4_res}
	\end{subfigure}
	\caption{Resulting viewpoints and 3D visualization of the target building.}
	\label{fig:bulidings_res}
\end{figure*}

In our experiments, we compared our method with the random sampling method \cite{jing2016sampling} using four different building models and sensor parameters. Results show that our method requires fewer viewpoints and achieves better coverage than random sampling.

\subsection{Setup}

Fig. \ref{fig:bulidings} shows four different target building models used to verify the effectiveness of our proposed method. In the comparative experiments, the preset target coverage is 100\%, i.e., $\delta=1$. Additionally, we set up several sensor parameters for comparative testing during the experiment, i.e., we use the maximum field of depth (FOD) of 30 meters and 40 meters, and the maximum field of view (FOV) of \SI{70}{\degree} and \SI{80}{\degree} respectively for view planning experiments. 

Candidate viewpoints are generated using both the clustering-based method proposed in this paper as well as the random sampling method, followed by the application of the same optimization algorithm. Both methods are executed 10 times with varying parameters in the experiments to illustrate the statistical comparison results.

\subsection{Results}

Fig. \ref{fig:result2} shows the relationship between the number of sampled candidate viewpoints and the uncovered area ratio, comparing two methods with varied sensor parameters settings. The curves are results fitted from experimental data, indicating that the proposed method consistently achieves better coverage and faster convergence by effectively utilizing the geometric information of the target building through spectral clustering. Conversely, the randomness of the random sampling method and hidden corners in the target structure may extend the time required for view planning.

\begin{table}[ht]
	\centering
	\setlength{\tabcolsep}{4pt}
	\caption{Comparison of view planning results}
	\begin{subtable}{0.5\textwidth}
		\caption{Number of resultant viewpoints}
		\label{tab:main_results_number}
		\begin{tabular}{@{}ccccccccc@{}}
			\toprule
			\multirow{2}{*}{params} & \multicolumn{2}{c}{\parbox{1.3cm}{\centering building1}} & \multicolumn{2}{c}{\parbox{1.3cm}{\centering building2}} & \multicolumn{2}{c}{\parbox{1.3cm}{\centering building3}} & \multicolumn{2}{c}{\parbox{1.3cm}{\centering building4}} \\ 
			\cmidrule(r){2-3} \cmidrule(r){4-5} \cmidrule(r){6-7} \cmidrule(r){8-9}
			& \parbox{0.65cm}{\centering Base-\\line} & \parbox{0.65cm}{\centering Pro-\\pose} & \parbox{0.65cm}{\centering Base-\\line} & \parbox{0.65cm}{\centering Pro-\\pose} & \parbox{0.65cm}{\centering Base-\\line} & \parbox{0.65cm}{\centering Pro-\\pose} & \parbox{0.65cm}{\centering Base-\\line} & \parbox{0.65cm}{\centering Pro-\\pose} \\
			\midrule
			(30, 80$^\circ$) & 69.0 & 52.7 & 132.3 & 101.9 & 110.7 & 78.3 & 101.6 & 82.6 \\
			(30, 70$^\circ$) & 85.2 & 64.1 & 160.4 & 123.1 & 136.7 & 95.6 & 120.5 & 95.8 \\
			(40, 80$^\circ$) & 45.2 & 35.2 & 88.3 & 71.9 & 67.9 & 63.3 & 66.9 & 57.5 \\
			(40, 70$^\circ$) & 54.9 & 42.3 & 103.6 & 86.0 & 88.8 & 72.8 & 79.0 & 64.9 \\
			\bottomrule
		\end{tabular}
	\end{subtable}

	\vspace{1em} % Add space between the subtables
	\begin{subtable}{0.5\textwidth}	
		\caption{Time efficiency comparison (unit: seconds)}
		\label{tab:main_results_time}
		\begin{tabular}{@{}ccccccccc@{}}
			\toprule
			\multirow{2}{*}{params} & \multicolumn{2}{c}{\parbox{1.3cm}{\centering building1}} & \multicolumn{2}{c}{\parbox{1.3cm}{\centering building2}} & \multicolumn{2}{c}{\parbox{1.3cm}{\centering building3}} & \multicolumn{2}{c}{\parbox{1.3cm}{\centering building4}} \\ 
			\cmidrule(r){2-3} \cmidrule(r){4-5} \cmidrule(r){6-7} \cmidrule(r){8-9}
			& \parbox{0.65cm}{\centering Base-\\line} & \parbox{0.65cm}{\centering Pro-\\pose} & \parbox{0.65cm}{\centering Base-\\line} & \parbox{0.65cm}{\centering Pro-\\pose} & \parbox{0.65cm}{\centering Base-\\line} & \parbox{0.65cm}{\centering Pro-\\pose} & \parbox{0.65cm}{\centering Base-\\line} & \parbox{0.65cm}{\centering Pro-\\pose} \\
			\midrule
			(30, 80$^\circ$) & 36.95 & 56.33 & 30.02 & 25.95 & 28.92 & 20.45 & 63.18 & 65.00 \\
			(30, 70$^\circ$) & 35.53 & 58.04 & 37.78 & 29.56 & 36.83 & 21.91 & 64.12 & 71.68 \\
			(40, 80$^\circ$) & 53.23 & 63.49 & 39.12 & 63.99 & 41.48 & 30.06 & 88.14 & 72.10 \\
			(40, 70$^\circ$) & 47.79 & 60.43 & 41.47 & 60.42 & 52.44 & 31.46 & 91.05 & 115.80 \\
			\bottomrule
		\end{tabular}
	\end{subtable}
	\label{tab:main_results}
\end{table}

As shown in Table \ref{tab:main_results_number}, for four different building models, the proposed method reduces the number of required viewpoints by 23.4\%, 20.4\%, 21.0\%, and 17.78\%, with an average reduction of 20.65\%, compared to the random sampling method. Under different target building models and sensor parameters, the time consumption comparison between the two methods shows varying advantages, as shown in Table \ref{tab:main_results_time}. In all experimental groups, both methods achieved 100\% coverage. 

Based on these results, we can infer that the proposed view planning method achieves higher coverage with fewer required viewpoints and does not result in a significant increase in computational costs, making it superior to existing methods. Fig. \ref{fig:bulidings_res} shows an example of the 3D visualization results from the proposed method for planning viewpoints.

\begin{table}[htbp]
	\centering
	\caption{Optimization results}
	\begin{tabular}{cccccc}
		\toprule
		\multirow{2}{*}{groups} & \multicolumn{2}{c}{viewpoints number} & \multicolumn{2}{c}{iteration number} \\
		\cmidrule(lr){2-3} \cmidrule(lr){4-5}
		& GA & GA-HH & GA & GA-HH \\
		\midrule
		1 & 49.0 & 49.0 & 119.0 & 45.8 \\
		2 & 48.0 & 48.0 & 47.0 & 35.8 \\
		3 & 48.0 & 48.0 & 17.6 & 18.4 \\
		4 & 53.0 & 53.0 & 45.0 & 33.8 \\
		5 & 50.0 & 50.0 & 88.4 & 71.0 \\
		6 & 52.0 & 52.0 & 82.0 & 42.4 \\
		7 & 54.4 & 54.0 & 68.6 & 49.2 \\
		8 & 53.0 & 53.0 & 42.4 & 26.4 \\
		9 & 52.0 & 52.0 & 68.8 & 48.4 \\
		10 & 54.0 & 54.0 & 43.4 & 42.6 \\
		averages & 51.34 & 51.30 & 62.22 & 41.38 \\
		\bottomrule
	\end{tabular}
	\label{tab:optimization_results}
\end{table}

To assess the effectiveness of the newly proposed genetic-based hyper-heuristic algorithm, we compared it with a commonly used combinatorial optimization algorithm, i.e., the genetic algorithm. 
Both algorithms are initialized with the same population size, and their crossover and mutation probabilities are adjusted respectively to achieve optimal performance. The genetic algorithm employed in the comparison has a similar high-level heuristic strategy to GA-HH. Based on the ten sets of experimental data under the first set of sensor parameters shown in Fig. \ref{fig:result2_1}, we obtained five optimization results for each optimization algorithm and calculated the average value as the result. 
As shown in Table \ref{tab:optimization_results}, the focus of the comparison was on the number of optimized viewpoints and the number of iterations required to reach the optimal solution. The results indicate that GA-HH converges more rapidly and is less likely to fall into local optima compared to the traditional genetic algorithm.

\section{Conclusions}

This paper proposes a new clustering-based view planning method for drone 3D visual coverage aimed at building surface inspection. Candidate viewpoints are efficiently generated based on spectral clustering and the local potential field method, and the problem is converted into a Set Covering Problem by calculating the visibility matrix, which is then solved with a genetic-based hyper-heuristic algorithm. 

Experimental results demonstrate that the proposed method generates higher quality candidate viewpoints compared to previous approaches, achieving greater coverage with fewer viewpoints and enabling more effective planning. In future work, we plan to explore coverage path planning based on this view planning method, aiming to optimize for the shortest drone inspection path or the least time, thereby enhancing the efficiency of building surface detection.

\addtolength{\textheight}{-1cm}   % This command serves to balance the column lengths
                                  % on the last page of the document manually. It shortens
                                  % the textheight of the last page by a suitable amount.
                                  % This command does not take effect until the next page
                                  % so it should come on the page before the last. Make
                                  % sure that you do not shorten the textheight too much.

%%%%%%%%%%%%%%%%%%%%%%%%%%%%%%%%%%%%%%%%%%%%%%%%%%%%%%%%%%%%%%%%%%%%%%%%%%%%%%%%

\bibliographystyle{IEEEtran}
\bibliography{References}

% Generated by IEEEtran.bst, version: 1.14 (2015/08/26)
\begin{thebibliography}{10}
\providecommand{\url}[1]{#1}
\csname url@samestyle\endcsname
\providecommand{\newblock}{\relax}
\providecommand{\bibinfo}[2]{#2}
\providecommand{\BIBentrySTDinterwordspacing}{\spaceskip=0pt\relax}
\providecommand{\BIBentryALTinterwordstretchfactor}{4}
\providecommand{\BIBentryALTinterwordspacing}{\spaceskip=\fontdimen2\font plus
\BIBentryALTinterwordstretchfactor\fontdimen3\font minus
  \fontdimen4\font\relax}
\providecommand{\BIBforeignlanguage}[2]{{%
\expandafter\ifx\csname l@#1\endcsname\relax
\typeout{** WARNING: IEEEtran.bst: No hyphenation pattern has been}%
\typeout{** loaded for the language `#1'. Using the pattern for}%
\typeout{** the default language instead.}%
\else
\language=\csname l@#1\endcsname
\fi
#2}}
\providecommand{\BIBdecl}{\relax}
\BIBdecl

\bibitem{jing2016sampling}
W.~Jing, J.~Polden, W.~Lin, and K.~Shimada, ``Sampling-based view planning for
  3d visual coverage task with unmanned aerial vehicle,'' in \emph{2016
  IEEE/RSJ International Conference on Intelligent Robots and Systems
  (IROS)}.\hskip 1em plus 0.5em minus 0.4em\relax IEEE, 2016, pp. 1808--1815.

\bibitem{TAN2021103881}
Y.~Tan, S.~Li, H.~Liu, P.~Chen, and Z.~Zhou, ``Automatic inspection data
  collection of building surface based on bim and uav,'' \emph{Automation in
  Construction}, vol. 131, p. 103881, 2021.

\bibitem{liu2024uav}
H.~Liu, Y.~P. Tsang, C.~K. Lee, and C.~H. Wu, ``Uav trajectory planning via
  viewpoint resampling for autonomous remote inspection of industrial
  facilities,'' \emph{IEEE Transactions on Industrial Informatics}, 2024.

\bibitem{9134859}
M.~Lauri, J.~Pajarinen, J.~Peters, and S.~Frintrop, ``Multi-sensor
  next-best-view planning as matroid-constrained submodular maximization,''
  \emph{IEEE Robotics and Automation Letters}, vol.~5, no.~4, pp. 5323--5330,
  2020.

\bibitem{koutecky2016sensor}
T.~Kouteck{\`y}, D.~Palou{\v{s}}ek, and J.~Brandejs, ``Sensor planning system
  for fringe projection scanning of sheet metal parts,'' \emph{Measurement},
  vol.~94, pp. 60--70, 2016.

\bibitem{scott2003view}
W.~R. Scott, G.~Roth, and J.-F. Rivest, ``View planning for automated
  three-dimensional object reconstruction and inspection,'' \emph{ACM Computing
  Surveys (CSUR)}, vol.~35, no.~1, pp. 64--96, 2003.

\bibitem{chen2011active}
S.~Chen, Y.~Li, and N.~M. Kwok, ``Active vision in robotic systems: A survey of
  recent developments,'' \emph{The International Journal of Robotics Research},
  vol.~30, no.~11, pp. 1343--1377, 2011.

\bibitem{jing2018model}
W.~Jing and K.~Shimada, ``Model-based view planning for building inspection and
  surveillance using voxel dilation, medial objects, and random-key genetic
  algorithm,'' \emph{Journal of Computational Design and Engineering}, vol.~5,
  no.~3, pp. 337--347, 2018.

\bibitem{maboudi2023review}
M.~Maboudi, M.~Homaei, S.~Song, S.~Malihi, M.~Saadatseresht, and M.~Gerke, ``A
  review on viewpoints and path planning for uav-based 3d reconstruction,''
  \emph{IEEE Journal of Selected Topics in Applied Earth Observations and
  Remote Sensing}, 2023.

\bibitem{9695293}
M.~Naazare, F.~G. Rosas, and D.~Schulz, ``Online next-best-view planner for
  3d-exploration and inspection with a mobile manipulator robot,'' \emph{IEEE
  Robotics and Automation Letters}, vol.~7, no.~2, pp. 3779--3786, 2022.

\bibitem{scott2009model}
W.~R. Scott, ``Model-based view planning,'' \emph{Machine Vision and
  Applications}, vol.~20, no.~1, pp. 47--69, 2009.

\bibitem{choi2018three}
Y.~Choi, Y.~Choi, S.~Briceno, and D.~N. Mavris, ``Three-dimensional uas
  trajectory optimization for remote sensing in an irregular terrain
  environment,'' in \emph{2018 International Conference on Unmanned Aircraft
  Systems (ICUAS)}.\hskip 1em plus 0.5em minus 0.4em\relax IEEE, 2018, pp.
  1101--1108.

\bibitem{wang2023high}
Y.~Wang, T.~Peng, W.~Wang, and M.~Luo, ``High-efficient view planning for
  surface inspection based on parallel deep reinforcement learning,''
  \emph{Advanced Engineering Informatics}, vol.~55, p. 101849, 2023.

\bibitem{weerasena2022design}
L.~Weerasena, A.~Ebiefung, and A.~Skjellum, ``Design of a heuristic algorithm
  for the generalized multi-objective set covering problem,''
  \emph{Computational Optimization and Applications}, vol.~82, no.~3, pp.
  717--751, 2022.

\bibitem{dokeroglu2023hyper}
T.~Dokeroglu, T.~Kucukyilmaz, and E.-G. Talbi, ``Hyper-heuristics: A survey and
  taxonomy,'' \emph{Computers \& Industrial Engineering}, p. 109815, 2023.

\bibitem{rivera2023aco}
G.~Rivera, L.~Cruz-Reyes, E.~Fernandez, C.~Gomez-Santillan, N.~Rangel-Valdez,
  and C.~A.~C. Coello, ``An aco-based hyper-heuristic for sequencing
  many-objective evolutionary algorithms that consider different ways to
  incorporate the dm's preferences,'' \emph{Swarm and Evolutionary
  Computation}, vol.~76, p. 101211, 2023.

\bibitem{zhang2021multitask}
S.~Zhang, Y.~Xu, and W.~Zhang, ``Multitask-oriented manufacturing service
  composition in an uncertain environment using a hyper-heuristic algorithm,''
  \emph{Journal of Manufacturing Systems}, vol.~60, pp. 138--151, 2021.

\bibitem{ferreira2015ant}
A.~S. Ferreira, A.~Pozo, R.~A. Gon{\c{c}}alves \emph{et~al.}, ``An ant colony
  based hyper-heuristic approach for the set covering problem,'' \emph{Advances
  in Distributed Computing and Artificial Intelligence Journal}, vol.~4, no.~1,
  pp. 1--21, 2015.

\bibitem{ng2001spectral}
A.~Ng, M.~Jordan, and Y.~Weiss, ``On spectral clustering: Analysis and an
  algorithm,'' \emph{Advances in neural information processing systems},
  vol.~14, 2001.

\bibitem{he2022diagnosis}
T.~He, S.~Zhu, H.~Wang, J.~Wang, and T.~Qing, ``The diagnosis of satellite
  flywheel bearing cage fault based on two-step clustering of multiple acoustic
  parameters,'' \emph{Measurement}, vol. 201, p. 111683, 2022.

\bibitem{shi2000normalized}
J.~Shi and J.~Malik, ``Normalized cuts and image segmentation,'' \emph{IEEE
  Transactions on pattern analysis and machine intelligence}, vol.~22, no.~8,
  pp. 888--905, 2000.

\bibitem{tian2021overall}
Y.~Tian, X.~Zhu, D.~Meng, X.~Wang, and B.~Liang, ``An overall configuration
  planning method of continuum hyper-redundant manipulators based on improved
  artificial potential field method,'' \emph{IEEE Robotics and Automation
  Letters}, vol.~6, no.~3, pp. 4867--4874, 2021.

\bibitem{moller2005fast}
T.~M{\"o}ller and B.~Trumbore, ``Fast, minimum storage ray/triangle
  intersection,'' in \emph{ACM SIGGRAPH 2005 Courses}, 2005, p.~7.

\bibitem{revuelta2021optimization}
E.~C. Revuelta, M.-J. Ch{\'a}vez, J.~A.~B. Vera, Y.~F. Rodr{\'\i}guez, and
  M.~C. S{\'a}nchez, ``Optimization of laser scanner positioning networks for
  architectural surveys through the design of genetic algorithms,''
  \emph{Measurement}, vol. 174, p. 108898, 2021.

\bibitem{wang2020improved}
K.~Wang, Y.~Gong, Y.~Peng, C.~Hu, and N.~Chen, ``An improved fusion crossover
  genetic algorithm for a time-weighted maximal covering location problem for
  sensor siting under satellite-borne monitoring,'' \emph{Computers \&
  Geosciences}, vol. 136, p. 104406, 2020.

\end{thebibliography}

\end{document}